\documentclass[letterpaper, 10 pt, journal, twoside]{ieeetran} 

\IEEEoverridecommandlockouts                              

\usepackage{graphicx} 
\usepackage{amsmath} 
\usepackage{amssymb}  
\usepackage{xfrac}
\usepackage{gensymb}
\usepackage{subfig}

\newcommand\bigzero{\makebox(0,0){\text{\huge0}}}

\title{Hybrid Open-Loop Closed-Loop Control of Coupled Human-Robot Balance During Assisted Stance Transition with Extra Robotic Legs*}

\author{Daniel J. Gonzalez, \IEEEmembership{Student Member,~IEEE}$^{1}$, and H. Harry Asada, \IEEEmembership{Member,~IEEE}$^{1}$
\thanks{Manuscript received: September, 10, 2018; Revised December, 8, 2018; Accepted January, 4, 2019.}
\thanks{This paper was recommended for publication by Editor Pietro Valdastri upon evaluation of the Associate Editor and Reviewers' comments. *This work was supported by National Robotics Initiative NRI-1637969 through US Department of Energy DE-EM0004484}
\thanks{$^{1}$The authors are with the d'Arbeloff Laboratory for Information Systems and Technology in the Department of Mechanical Engineering,
        Massachusetts Institute of Technology, Cambridge, MA 02139, USA. Email: 
        {\tt\small \{dgonz, asada\}@mit.edu}}%
\thanks{Digital Object Identifier (DOI): see top of this page.}
}

\begin{document}

\maketitle

\begin{abstract}
A new approach to the human-robot shared control of the Extra Robotic Legs (XRL) wearable augmentation system is presented. The XRL system consists of two extra legs that bear the entirety of its backpack payload, as well as some of the human operator's weight.
The XRL System must support its own balance and assist the operator stably while allowing them to move in selected directions.
In some directions of the task space the XRL must constrain the human motion with position feedback for balance, while in other directions the XRL must have no position feedback, so that the human can move freely. Here, we present Hybrid Open-Loop / Closed-Loop Control Architecture for mixing the two control modes in a systematic manner. The system is reduced to individual joint feedback control that is simple to implement and reliable against failure. The method is applied to the XRL system that assists a human in conducting a nuclear waste decommissioning task. A prototype XRL system has been developed and demonstrated with a simulated human performing the transition from standing to crawling and back again while coupled to the prototype XRL system. 
\end{abstract}



\section{Introduction}\label{secIntro}
\IEEEPARstart{P}erforming work tasks near the ground while carrying or wearing heavy equipment can be immensely fatiguing, especially when the task involves frequently transitioning from one location or position to another. For example, nuclear decommissioning workers for the United States Department of Energy (DOE) must carry 13.6-kilogram (30-pound) Aluminum SCBA systems in addition to their HAZMAT Level A suits during an emergency response mission, and exhaustion usually sets in before the tank can be depleted of its 30 minutes of air \cite{NIOSH1985}. Working and maneuvering with such a heavy payload for extended periods of time can lead to repetitive back strain injuries, which can ultimately cost employers. According to the US Bureau of Labor Statistics, there were over 190,000 workplace injuries in manufacturing sectors, and 50,000 injuries in agriculture in 2014 \cite{BureauofLaborStatistics2015a}. In 2010, the average civilian worker's compensation claim due to a back injury in the USA was between \$40,000 and \$80,000 \cite{CalgaryRegionalHealthAuthority2010}. 
\begin{figure}[t!]
	\centering
	\includegraphics[width=0.55\linewidth]{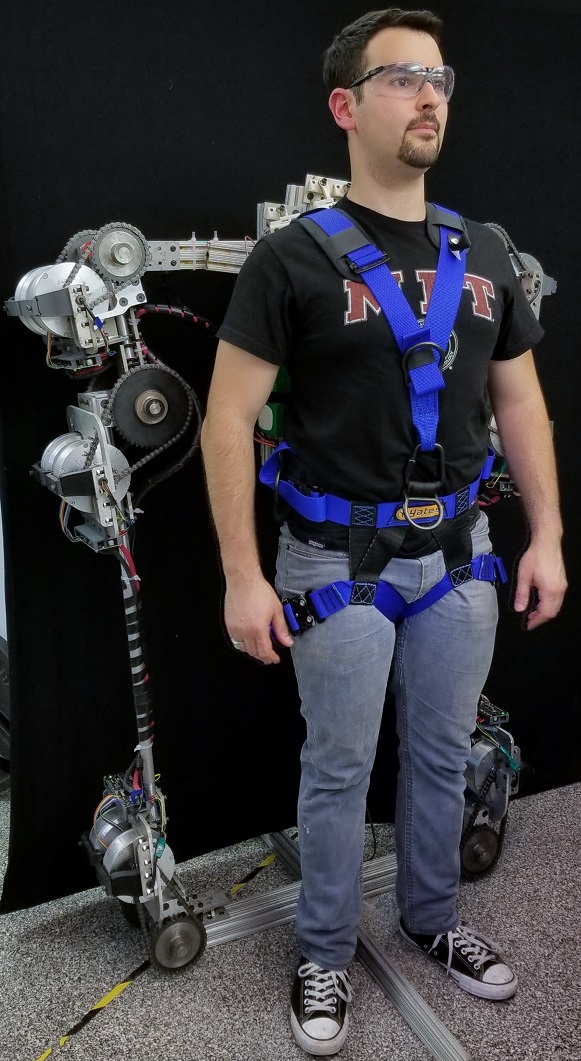}
	\caption{The Prototype Extra Robotic Legs (XRL) System being worn by an operator.}
	\label{figXRLWorn}
\end{figure}

\begin{figure}[t]
	\centering
	\includegraphics[width=\linewidth]{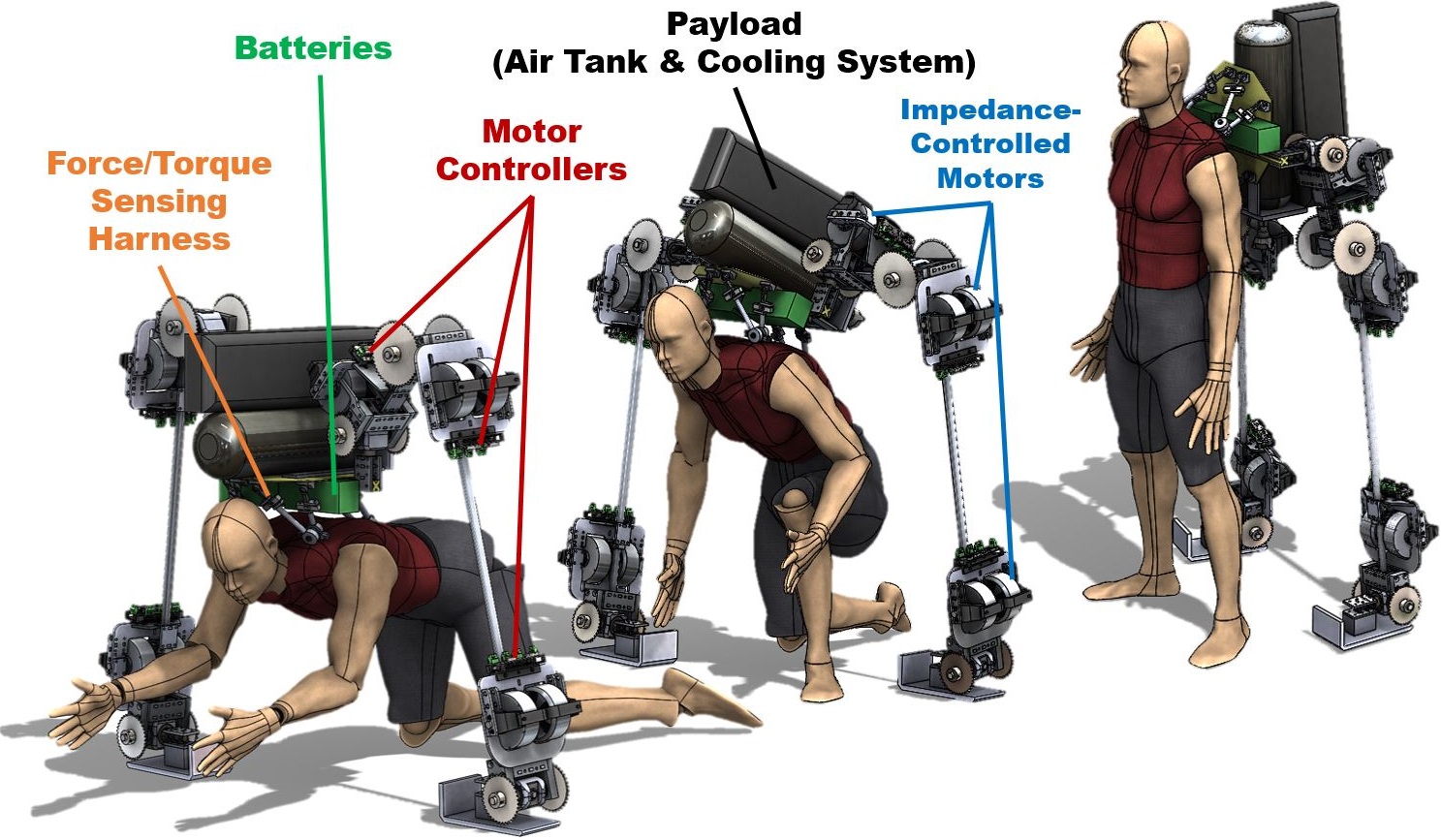}
	\caption{The Extra Robotic Legs (XRL) System throughout the transition between the standing and crawling configurations, and a description of the key system components.}
	\label{figXRLSystem}
\end{figure}

The Extra Robotic Limbs (XRL) System aims to alleviate the physical issues associated with ground work and reduce the monetary costs to employers by supporting the weight of an equipment payload both while standing and while crawling. The XRL System (See Fig. \ref{figXRLWorn}) is essentially a backpack with its own legs, enabling a human operator to walk around, climb stairs, and crawl on the ground completely unhindered by their heavy payload (See Fig. \ref{figXRLSystem}). Previous work on Supernumerary Robotic Limbs from our laboratory focused on only the ground task \cite{Kurek2017}. Our first publication on the XRL System describes the design driven by the task requirements of energy efficiency, proprioception for safe human interaction, and the ability to bear heavy loads \cite{Gonzalez2018}. We minimize the maximum required torque by exploiting the closed kinematic chain to bear the heavy loads present during the transition from standing to squatting. Future plans include footstep planning based on observation of the operator motion, analysis of dynamic balance control while walking, and synchronizing the gait cycle with that of the human operator for efficient and comfortable walking.


We now aim to control the balance and fine interaction between the human and the XRL System during its most heavily loaded operating condition: the transition from standing to crawling. During this task there is both a requirement of maintaining balance and a requirement of applying an upward assistive force to the human operator while allowing them to dictate the pace of stance transition through physical interaction alone. This interaction is a type of shared control between a robot and a human \cite{Mulder2015}. The robot constrains the human body for balancing and stability, while the human determines the time of the transition and the pace of its execution. It is natural to divide the task coordinate space into the one controlled by the robot and another governed by the human. 
A different control mode must be assigned to each of the two subspaces in the task coordinate space, then converted to joint subspaces for physical realization. 

There are many prior works on related topics. Hybrid position-force control allows us to blend position control and force control in orthogonal subspaces \cite{Raibert1981}\cite{Fisher1992}. Impedance and admittance control also allow us to assign different stiffness and damping parameters to individual axes \cite{Hogan1984}\cite{Seraji1994}\cite{Schimmels1992}. By varying the robot Cartesian impedance relative to endpoint velocity, 
while it interacts with the human operator in performing certain tasks, the task accuracy, execution time, and human sense of comfort can be improved \cite{Ficuciello2015}. While these control methods are powerful in creating and synthesizing desired behaviors in the task space, there are other requirements and conditions that must be met in order to apply these control methods to physically coupled human-robot systems, such as the XRL system. 

As the XRL is directly attached to a human body, safety is a primary concern. Simple, robust, and failsafe design is a critical requirement for control design and system architecture. Even with some of the cables and communication lines are disabled, the system must be able to maintain some minimum functions. Complex multivariate control, which entails the transmission of numerous signals through various bus lines and cables, tends to be vulnerable. Simple independent joint servos with local feedback are preferred in industrial settings. From the practical implementation and maintenance viewpoint, simple individual joint feedback systems are advantageous. Prior work on distributing the control task to the joint servo controllers includes \cite{Albu-Schaffer2002}.

This paper aims to develop a control algorithm and system architecture that can realize a dual control mode in a task space that can failsafe to individual joint controls in the case of a communications issue, but continuously operates with fully-coupled joint control otherwise. The joint control systems are reliable and easy to implement. Furthermore, the system can be made failsafe. In the following, Section \ref{secTask} describes the task and control requirements, Section \ref{secHOCCA} describes the principle of the new algorithm, followed by the application of the algorithm to the XRL performing squatting and transition between standing and squatting in Section \ref{secApp}. Section \ref{secImplExp} describes the implementation of our control methods on the XRL Prototype, and Section \ref{secConclusion} provides a conclusion and an outline of our plan for future work.

\section{Task Description and Requirements}\label{secTask}
The XRL System is a pair of fully-actuated articulated legs with 6 degrees of freedom each whose hips are coupled to a base ``torso'' (See Fig. \ref{figXRLSystem}). The backpack payload is mounted rigidly to the torso, allowing the legs to balance and bear the weight. The torso is also coupled to the human operator via a 6-DOF Force-Torque sensing harness\cite{Ruiz2017}, which is used to measure the full interaction wrench between the XRL torso and the operator. 

While idly standing, the human operator may bear their own weight while the XRL System stands behind them and carries only their backpack load. To reach the ground and perform some work task, the human operator must stoop, crouch down, and ultimately transition to a crawling posture. Once crawling, the operator must be able to use both hands freely to perform the work task, which requires significant support of their torso. See Fig. \ref{figXRLSystem}.
\begin{figure}[t]
	\centering
	\includegraphics[width=0.9\linewidth]{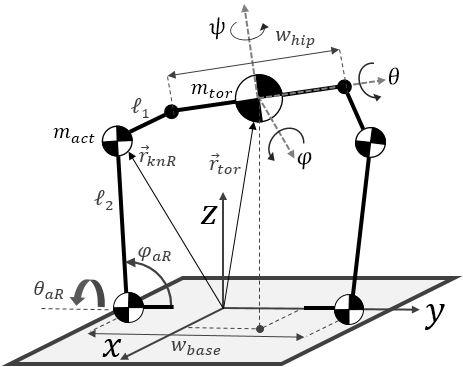}
	\caption{Geometric description and mass distribution Model for the XRL. The robot has 12 DOFs in total. Each of the two legs has three 2-DOF joint modules: one at the hip, one at the knee, and one at the ankle. Each joint module can rotate in a sagittal $\theta$ and frontal $\varphi$ dimension relative to its base. See Section  \ref{secApp} for details.}
	\label{figCOM}
\end{figure}

During this transition to the ground and while crawling on the ground, the human operator must be allowed to change their vertical position $Z$ and body orientation $\theta$ significantly. See Fig. \ref{figCOM} for coordinate system details. This transition is particularly difficult because the robot must pass through its least kinematically efficient configuration, and thus apply its maximum operating torque, as noted in our previous design paper. Proper balance and behavior during this most sensitive of motions is key to ensuring the stability of the XRL System and the safety and comfort of the human operator.

Safety must be taken into consideration in the design of the control architecture because the robot will be interacting closely with its human operator. Perhaps the most widely known examples of whole body balance control failure are those of the humanoid robots that competed in the DARPA Robotics Challenge \cite{Guizzo2015} which suffered a variety of issues \cite{Atkeson2015}. Both passive and active exoskeletons are beginning to see adoption in industry worldwide, and regulations have been set in place to ensure the safety of their operators \cite{VanderVorm2015}. One strategy for increasing the reliability of a control system is to ensure that the low-level hardware and control firmware can operate and maintain stability independent of the higher-level software components \cite{Roderick2007}. We require for the XRL a failsafe design in the overall stability control that is distributed to the lowest level in the case of communications latency, delay or failure. 

Note that for this work, we investigate only the situation where the feet are firmly planted on the ground and do not move. Current and future work seeks to address the problem of stability and synchronization of gaits while walking and crawling. 

The control design of the XRL System for the crawl transition task was driven by the following specific requirements. The XRL System must:
\begin{itemize}
	\item  Keep its balance. Specifically, the XRL System must regulate its center of gravity towards the center of its support polygon along the horizontal $X$ and $Y$ directions.
	\item Maintain stability of sensitive rotation movements to prevent unwanted twisting motions. These include roll $\varphi$ and yaw $\psi$ rotations of the torso.
	\item Apply an assistive upward $Z$ force  to the operator both while in the crawling state and while standing up or crouching down.
 \item Allow the human to govern the transition process, control the pace of the transition through natural force interaction alone.
 \item Allow the human to change their pitch angle $\theta$ during the transition between standing and crawling.
 \item Remain standing and balanced in the case of a communications delay or failure between the centralized controller and the individual joint servo controllers. 
\end{itemize}

\section{Hybrid Open-Loop Closed-Loop Control Architecture (HOCCA)}\label{secHOCCA}
\subsection{Concept}
In shared control, some axes are controlled by autonomy based on reference inputs generated by the machine, while other axes are controlled based on exogenous inputs, which are typically human inputs. For the wearable XRL robot, the latter are the human posture, the height and pitch of the upper body, indicating the human intention of standing up, kneeing down and crawling as well as the pace of posture change. The task space is divided into two subspaces orthogonal to each other: one is a robot-controlled subspace and the other a human-controlled subspace. In the robot-controlled subspace, closed-loop control of positional variables is formed with reference inputs provided by the robot. In the human-controlled subspace, the closed-loop control of the robot-generated references must be eliminated, and the robot must not impede the human motion, but conform to it. This entails that the task space be divided into the subspace with closed-loop control of robot-generated references and the one where the closed-loop position control is eliminated. This task space division of control mode can be translated and decomposed into joint control systems. As a result, each joint control system is a mixture of closed-loop control and open-loop control of the robot-generated references.
\begin{figure}
\centering
\includegraphics[width=\linewidth]{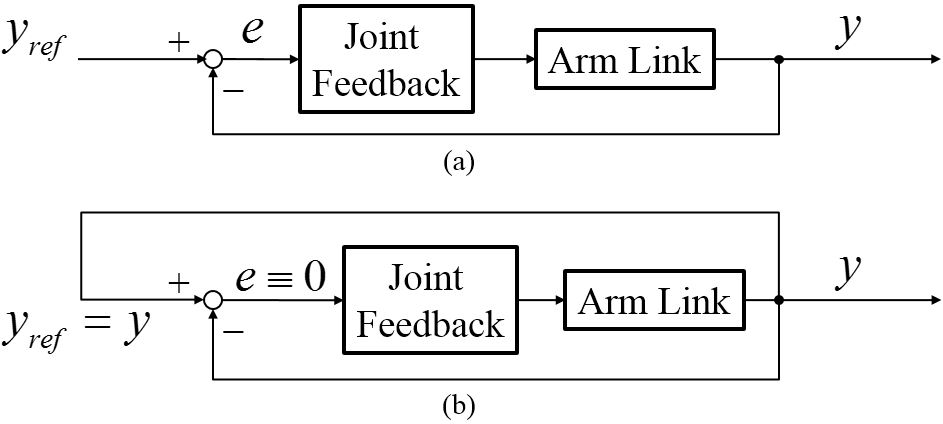}
\caption{Single joint control system: (a) Standard joint feedback system, (b) Virtually disconnected feedback system}
\label{figSingle}
\end{figure}

Fig. \ref{figSingle}-(a) shows a standard single joint control system with position feedback. Simple Proportional and Derivative (PD) control can stabilize not only the individual joint control system, but also the coupled multi-axis control system if the individual joints are connected to the entire robot system. Our objective is to make this primitive individual joint control system be hybrid of closed-loop and open-loop controls. The feedback loop must exist for the robot controlled subspace. Therefore, we cannot physically disconnect the feedback loop for the human controlled subspace. Instead, we virtually disconnect the feedback loop. As shown in Fig. \ref{figSingle}-(b), consider that the reference input to the position control system is replaced by the output of the system: $y_{ref}\equiv y$. This results in zero feedback error: $\epsilon\equiv 0$. Therefore, the feedback loop is virtually disconnected. When the open-loop and closed loop control modes are converted from the task space to the joint space, each joint control system must serve partially as a closed-loop and partially as an open-loop system. This can be achieved by manipulating the reference inputs. The component of the reference input associated with the open-loop mode in the task space is set to the corresponding component of the current output, which does not generate any feedback error signal in that subspace. Therefore, the joint controller would not interfere with the human motion.

Let $p \in V^6$ be a 6-dimensional vector\footnote{The three Euler angles do not span a vector space, because the rotation matrices associated with them do not commute. Here, we assume small angular displacements, which are vectorial quantities.}, representing the position and orientation (pose) of a system's end-effector in a task space. The vector space $V^6$ is divided into a closed-loop control subspace $V_C\subset V^6$ and an open-loop control subspace $V_O\subset V^6$. It is assumed that the direct sum of these subspaces span the entire vector space:
\begin{equation}
V_C\oplus V_O =V^6 
\end{equation}

Let $p_C\in V_C$ be a pose of the end-effector in the closed-loop subspace, and $p_O\in V_O$ be a pose in the open-loop subspace. The sum of these two, $p=p_C+p_O$, is in $V^6$. Let $S\in\Re ^6$ be the projection matrix that projects any vector in $V^6$ onto the open-loop subspace $V_O$. Namely,
\begin{equation}
 p_O = S \cdot p
\end{equation}
	  
Let $S^\perp$ be the projection matrix that is orthogonal complement to $S$. Then, 
\begin{equation}
S^\perp=I-S
\end{equation}
and
\begin{equation}
p_C = S^\perp\cdot p
\end{equation}

\subsection{Overall Control Architecture}
In the Hybrid Open-loop Closed-loop Control Architecture, the positional reference input $p_{ref}$ is constructed with two vectors in the closed-loop and open-loop subspaces, respectively. For the closed-loop subspace, $p_C\in V_C$ is a positional reference generated by the robot for closed-loop position control. For the open-loop subspace, $p_O\in V_O$ is the current output of the system in the open-loop subspace. Adding these two, we construct the positional reference input as
\begin{equation}
p_{ref}=p_C+p_O
\end{equation}
Note that
\begin{equation}
p_C = S^\perp \cdot p_{ref},~~p_O=S\cdot p_{ref}
\end{equation}

\begin{figure}[t]
	\includegraphics[width=\linewidth]{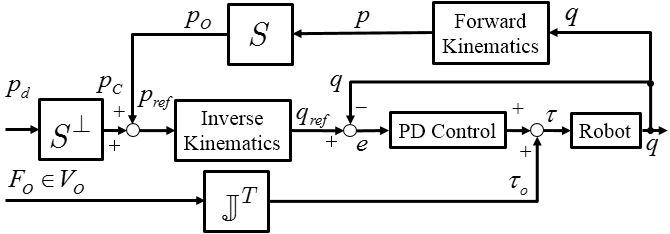}
	\caption{Block diagram of Hybrid Open-loop / Closed-loop Control Architecture}
	\label{figHOCCA}
\end{figure}

To resolve this reference command in the task space to the ones of individual joint feedback systems, we solve the inverse kinematics problem of the system. 
\begin{equation}
q_{ref}=IK\left(p_{ref}\right)
\end{equation}
where $q_{ref}\in \Re^n$ is the solution of the inverse kinematics, $n$ is the number of reference inputs in the joint coordinate system, and $IK()$ is the global inverse kinematics function relating $q$ to $p$. Fig. \ref{figHOCCA} shows the block diagram of the overall system.

In the case of the XRL System, the robot must support the human body by bearing a fraction of the body weight during the transition between standing and crawling. An appropriate force and moment must be generated at the individual joints to support the body. Such force and moment can be generated in the open-loop subspace as feedforward inputs to individual joint control systems. 

Let $F_O \in V_O$ be a force and moment (wrench) in the open-loop subspace. This force and moment in the task space can be distributed over the individual joints as:
\begin{equation}
\tau_O=\mathbb{J}^T F_O
\end{equation}
where $\tau_O \in \Re^n$ is a vector of joint torques to be generated at the individual joints and $\mathbb{J}$ is the Jacobian matrix relating joint velocities to endpoint velocities. This can be implemented as a feedforward input to each joint control system, as shown in Fig. \ref{figHOCCA}.
 
\subsection{Failsafe Joint Distributed Task Space Control}
Given the overall system architecture, we now introduce a controller defined in the task space and its distribution to the individual joint controllers. It may be favorable for the servo control loop to run at a low level without requiring high speed communication to a centralized controller that is generating torque commands. A slow transmission frequency, a time delay, or a total loss in communication would not be catastrophic and still allow the distributed portions of the controller to maintain the overall system's stability. 

We define a task space PD controller with an open-loop force term. 
\begin{equation}\label{ForceImpedance}
F=K_p\left(p_{ref}-p\right)+B_p\left(\dot{p}_{ref}-\dot{p}\right)+F_O
\end{equation}
where $K_p$ and $B_p$ are the constant  $6\times6$ stiffness and damping matrices regulating $p$ and $F$ is the total wrench to apply in task space.

To generate the joint actuator torques $\tau$ required to apply the desired task space wrench, we use the Jacobian in the following expression:
\begin{equation}\label{JtransposeF}
\tau=\mathbb{J}^TF
\end{equation}
We wish to have a final controller in a form we can send to our motor controllers:
\begin{equation}\label{JointLaw}
\tau = K_q(q_{ref}-q)+B_q(\dot{q}_{ref}-\dot{q})+\tau_0
\end{equation}
where $K_q$ and $B_q$ are, respectively,  $n\times n$ stiffness and damping matrices in the joint space. Our goal is to obtain $(K_q,B_q)$ that make our robot exhibit the desired Cartesian stiffness and damping $(K_p,B_p)$. We assume that $(K_q,B_q)$ are local and constant in the vicinity of the current configuration. We begin by taking partial derivatives of equations \eqref{JtransposeF} and \eqref{JointLaw} with respect to $q$:
\begin{equation}
\frac{\partial \tau}{\partial q} = -K_q = \left(\frac{d}{dq}\mathbb{J}^{T}\right) F + \mathbb{J}^{T}\left(\frac{\partial}{\partial q}F\right)
\end{equation}
From equation \eqref{ForceImpedance}, we obtain
\begin{equation}
K_q = \mathbb{J}^{T}K_p \mathbb{J} -\left(\frac{d}{dq}\mathbb{J}^{T}\right) F
\end{equation}
$\left(\frac{d}{dq}\mathbb{J}^{T}\right)$ is the Hessian tensor of order 3, expressed as the array of second partial derivative matrices. 
\begin{equation}\label{FullEqn}
K_q = \mathbb{J}^{T}K_p \mathbb{J} - \sum_{i=1}^{6} \mathbb{H}_i F_{i}
\end{equation}
where
\begin{equation}
\mathbb{H}_i =  \frac{\partial}{\partial q} \left(\frac{\partial p_i}{\partial q}\right)^{T} =  \begin{Bmatrix} \frac{\partial ^{2} p_i}{\partial q_j \partial q_k}\end{Bmatrix} , ~~~~i \in1:6
\end{equation}
This is a similar result as that computed in \cite{Hogan1990} and \cite{Chen1999}. Note that $\mathbb{J}$, $\mathbb{H}$, and $F$ are all functions of the joint configuration $q$. As the configuration $q$ changes, the central controller updates these variables. 

It is important to note that the above joint stiffness matrix $K_q$ is not necessarily  positive-definite. In particular, the XRL System consisting of two robotic limbs has a rectangular Jacobian matrix, $n > 6$, creating a nullspace. To stabilize the joint feedback controller, we add a nominal positive-definite joint stiffness $K_0$ to stabilize the nullspace.
\begin{equation}\label{KDGonz}
K_q = N K_0 + \mathbb{J}^{T}K_p \mathbb{J} - \sum_{i=1}^{6} \mathbb{H}_i F_{i}
\end{equation}
using the nullspace projection matrix
\begin{equation}
N = \left(I-\mathbb{J}^\dagger\mathbb{J}\right)
\end{equation}
where $\mathbb{J}^\dagger$ is the Jacobian pseudoinverse.
Adding this term we can find a positive-definite joint stiffness matrix while maintaining our desired Cartesian impedance. 

Similarly, differentiating \eqref{JtransposeF} with respect to $\dot{q}$ and adding a nominal diagonal joint damping $B_0$ gives us
\begin{equation}\label{BDGonz}
B_q = N B_0 + \mathbb{J}^{T}B_p \mathbb{J}
\end{equation}

\begin{figure}[t]
\includegraphics[width=\linewidth]{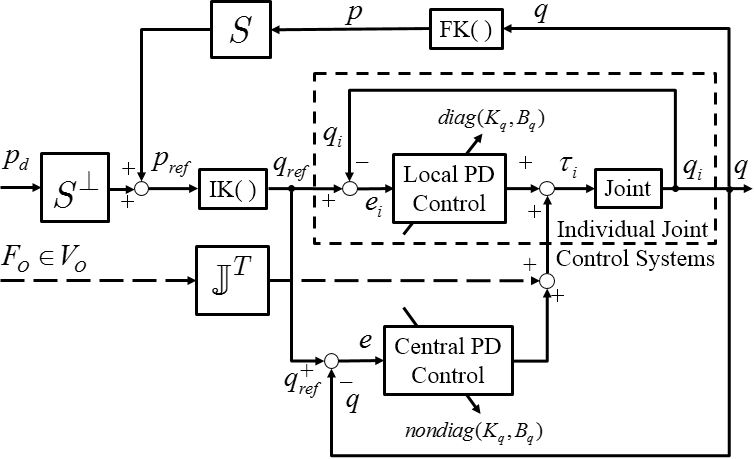}
\caption{Block Diagram of Joint Distributed Hybrid Open-Loop / Closed-Loop Control of Task Space Impedance. Note that $FK()$ and $IK()$ are the Forward Kinematics and Inverse Kinematics functions, respectively.
}
\label{figImpedance}
\end{figure}
In order to distribute this controller among the $n$ individual joint controllers, the multivariate controller must be split into two components: the coupled and decoupled control components. The diagonal terms of $K_q$ and $B_q$ can be used for the $n$ independent and decoupled joint controllers because $K_q$ and $B_q$ are guaranteed to be positive definite. We can then compute the non-diagonal cross-coupled multivariate terms in the centralized controller and feed the resultant torque to each joint controller along with $\tau_O$. See Fig. \ref{figImpedance} for a block diagram. The advantage of this control architecture is that the independent joint controllers can still operate stably with the diagonal terms alone although the central controller is unable to deliver the torque of the cross-coupled terms due to communications latency, delay or failure. 

\section{Application to the XRL System}\label{secApp}
We will now apply Hybrid Open-Loop / Closed-Loop Control to the XRL System. In addition to the kinematic model, the Jacobian $\mathbb{J}$, the open-loop projection matrix $S$, the task-space position reference for closed loop control $p_C\in V_C$, $p_C=S^\perp \cdot p$, the task-space assistive force $F_O\in V_O$, and the task-space impedance $(K_p, B_p)$ must be obtained to implement the controller.

We require stable balance control of the XRL System, namely, the ability to maintain the 3-DOF Center of Mass (COM) over the center of its support polygon (the convex hull of its ground contact). We additionally wish to control the 3-DOF orientation of the robot's torso, which we define as Euler angles following a $z$-$x$-$y$ (3-1-2) rotation order to coincide with the robot geometry. The combined 6-DOF vector 
\begin{equation} \label{eqpdef}
p=\begin{bmatrix}
\vec{r}_{COM} \\
\vec{\Theta}_{torso}
\end{bmatrix}=
\begin{bmatrix}
x_{COM}& y_{COM}& z_{COM}& \varphi& \theta& \psi 
\end{bmatrix}^T
\end{equation}
describes this pose we wish to control.

The mass distribution of the XRL prototype is dominated by the payload, batteries, actuators, and transmission elements. See TABLE \ref{tableProps} for details. The proximal and distal leg links, of length $\ell_1$ and $\ell_2$ respectively, can be considered massless relative to the other components. Each actuator/transmission module weighs an average of $m_{act}$ and is located at the joint it controls: the ankle module is located at the ankle joint, the knee module at the knee joint, and the hip module at the hip. The hips are a distance $w_{hip}$ from each other along the torso, and the ankles are a distance $w_{base}$ from each other along the ground. Each of the two batteries weighs $m_{batt}$, and they are located at the center of the torso. The payload mass $m_{pl}$ is variable, but generally located at the center of the torso. 

\begin{table}[t]
	\renewcommand{\arraystretch}{1.3}
  	\centering
  	\begin{tabular}{|c|c|}
      \hline
      $m_{actuator}$ $(m_{act})$& 8.73 kg (19.25 lbs)\\
      $m_{battery}$ $(m_{batt})$& 0.9 kg (2 lbs)\\
      $m_{payload}$ $(m_{pl})$ & 20.4 kg (45 lbs) maximum\\
      $F_{assist}$ & 200 N (45 lbs-force) maximum\\
      $\ell_1$ & 0.2667 m (10.5 inches)\\
      $\ell_2$ & 1.0290 m (40.5 inches)\\
      $w_{hip}$ & 0.7747 m (30.5 inches)\\
      $w_{base}$ & 0.6922 m (27.25 inches)\\
      $\ell_{foot}$ & 0.254 m (10 inches)\\
      $w_{foot}$ & 0.127 m (5 inches)\\
      \hline
  	\end{tabular}
  	\caption{Mass, payload, and length properties of the XRL Prototype}
  	\label{tableProps}
\end{table}

We may consider the batteries, hip modules, and payload to be located on the same rigid link, and thus the total torso mass is $$m_{torso} = 2m_{act}+2m_{batt}+m_{pl}$$ and located at the center of the torso link.

The joint locations $\vec{r}_{ji}$ where $i = Left,~Right$ and $j = Hip,~Knee,~Ankle$ are defined as follows:
\begin{equation}
\vec{r}_{ankleRight}=
\begin{bmatrix}
    0\\
	-w_{base}/2\\
    0
	\end{bmatrix}
\end{equation}
and the left ankle vector points in the opposite direction, symmetric about the origin. Position vectors $\vec{r}_{torso}$, $\vec{r}_{kneeLeft}$, and $\vec{r}_{kneeRight}$ can be found from applying simple 3D forward kinematics using joint angle feedback. These names are abbreviated in the equations for ease of formatting.

The total center of mass location is
\begin{equation}\label{eqR}
	\vec{r}_{COM} = \frac{\vec{r}_{tor}m_{tor}+(\vec{r}_{knL}+\vec{r}_{knR}+\vec{r}_{anL}+\vec{r}_{anR})m_{act}}{m_{tor}+4m_{act}}
\end{equation}

The rotation matrix relating the torso angle with respect to the origin is
\begin{equation}
	\mathbb{R}_{torso}= 
    \mathbb{R}_{xy}(\varphi_{aR},\theta_{aR})
    \mathbb{R}_{xy}(\varphi_{kR},\theta_{kR})
    \mathbb{R}_{xy}(\varphi_{hR},\theta_{hR})
\end{equation}
which takes each of the 6 joint angles associated with the right leg as arguments. Note that subscripts $aR$, $kR$,and $hR$ refer to the ankle, knee, and hip, respectively, of the right leg. 

Decomposing $\mathbb{R}_{torso}$ to the 3-1-2 Euler sequence, we get
\begin{equation}
\begin{split}
\mathbb{R}_{torso} &= \mathbb{R}_y(\theta)\mathbb{R}_x(\varphi)\mathbb{R}_z(\psi)\\ &=\left[  \begin{matrix}
c_{\theta} c_{\psi} - s_{\theta} s_{\varphi} s_{\psi} & c_{\theta} s_{\psi} + s_{\theta} s_{\varphi} c_{\psi} & -s_{\theta} c_{\varphi}  \\
-s_{\psi} c_{\varphi} & c_{\psi} c_{\varphi} & s_{\varphi} \\
s_{\theta} c_{\psi} + c_{\theta} s_{\varphi} s_{\psi} & s_{\theta} s_{\psi} - c_{\theta} s_{\varphi} c_{\psi} & c_{\theta} c_{\varphi} \end{matrix} \right]
\end{split}
\end{equation}
where
\begin{equation}
	\vec{\Theta}_{torso}=
	\begin{bmatrix}\varphi\\ \theta\\ \psi\end{bmatrix} = 
    \begin{bmatrix}
    \arcsin(\mathbb{R}_{torso23})\\ 
    \arctan2(-\mathbb{R}_{torso13},\mathbb{R}_{torso33})\\ 
    \arctan2(-\mathbb{R}_{torso21},\mathbb{R}_{torso22})
    \end{bmatrix}
\label{eqTheta}
\end{equation}

Equations \eqref{eqR} and \eqref{eqTheta} provide us with the kinematic model of our system, from which we may obtain the Jacobian matrix $\mathbb{J}$.

In addition to balancing and maintaining stability, the XRL system must bear its own weight and apply a constant upward force to the operator, allowing them to angle themselves forward and backward while crouching down and standing up. We define the task-space force vector to include the gravitational compensation force and upward assistive force
\begin{equation}
F_O = \left(gm_{total}+F_{assist}\right)\hat{z}
\end{equation}
where $F_{assist}$ is the assistive upward force applied to the operator and
\begin{equation}
m_{total}=m_{torso}+4m_{actuator}
\end{equation}
is the total robot mass that is acted on by gravity $g$.

The space in which these are defined make up the open-loop projection matrix
\begin{equation}
S=
\left[
\begin{array}{cccccc}
	0& & & & & \\
     &0& & &\bigzero& \\
     & &1& & & \\
     & & &0& & \\
     &\bigzero&& &1& \\
     & & & & &0
\end{array}
\right]
\end{equation}
which provides a filter to isolate the requested open-loop force subspace $V_O$ including the operator/robot height $z$ and pitch angle $\theta$. The position controlled subspace $V_C$ can be isolated using the closed-loop projection matrix $S^\perp = \left[I-S\right]$ which is the orthogonal complement of $S$.

Note that we may maintain the joint distribution property of the aforementioned whole body impedance controller by treating the open-loop subspace of the current position as external inputs to generate $\vec{q}_{ref}$ for the position controllers. Before the global inverse kinematics are taken, these two vectors are added together. Their components are always orthogonal due to the selection matrices, so no actual addition of values takes place. After passing through the selection matrices they become
\begin{equation}
p_C=
\begin{bmatrix}
x_{COM}& y_{COM}& 0& \varphi_{torso}& 0& \psi_{torso} 
\end{bmatrix}^T
\end{equation}
and
\begin{equation}
p_O=
\begin{bmatrix}
0& 0& z_{operator}&0& \theta_{operator}& 0 
\end{bmatrix}^T
\end{equation}

The body-coordinate gains for the dimensions of these external inputs are also ensured to be 0 using the position selection matrix $S^\perp$. The result is that the force controller only acts in the space defined by $S$ and the position controller only acts along the space defined by $S^\perp$. These gains are then also converted to the joint level using the Jacobian transform from equations \eqref{KDGonz} and \eqref{BDGonz}.

The choice of our body-coordinate stiffness gains in the most sensitive directions is important to the ability to balance stably. We provide a deterministic calculation for the minimum required body stiffness along the $x$-axis to achieve balance of the XRL.

The maximum torque capable of being produced by one ankle is $\tau_{ank~max}$. If we consider $z_{COM}=z_{max}$ at the fully upright and centered configuration with two feet on the ground, then lean forward by some angle $\theta$
Solving back for $\theta$, we get the maximum allowable lean angle assuming the feet are always coupled to the ground
\begin{equation}
\theta_{max}=\arcsin{\frac{2\tau_{ank~max}}{m_{tot}gz_{max}}}
\end{equation}

As is noted in \cite{Gonzalez2018}, squatting is performed with the feet configured in the frontal plane. If we only lean forward until the center of mass lies at the edge of the foot support polygon located $w_{foot}/2$ away from the origin, we must remain within the following maximum lean angle to maintain balance and not tip over:
\begin{equation}
\theta_{edge} = \arctan{ \frac{\ell_{foot}}{2z_{max}}}
\end{equation}
We can use this to compute the minimum required torque at the support polygon edge
\begin{equation}
\tau_{ank~edge} = \frac{m_{tot}gz_{max}}{2}\sin{\theta_{edge}}
\end{equation}

If $\tau_{ank~edge}<\tau_{ank~max}$, then we are always able to return to equilibrium from a static initial condition at the support polygon edge. The minimum required individual ankle stiffnesses are then 
\begin{equation}
K_{ank~min}=\frac{\tau_{ank~edge}}{\theta_{edge}}~[Nm/rad]
\end{equation}
which in the body space becomes
\begin{equation}
K_{x~min}=\frac{2K_{ank~min}}{z_{max}^2}~[N/m]
\end{equation}

For example, if we consider $\tau_{ankle~max} = 115.6$ Nm, $m_{total}=m_{top} = 57.12$ kg with maximum payload and ignoring the ankles (which require no support), $z_{max}=\ell_1+\ell_2=1.2957$ meters as an approximation of the upper body center of mass, we get $\theta_{edge}=2.81^\circ=0.0490$ radians and a $\tau_{edge}=17.76$ Nm requiring an individual ankle stiffness of $K_{ank~min}=362.65$ Nm/rad, or a body stiffness $K_{x~min}=432$ N/m. While this demonstrates a theoretical lower limit, we include a factor of safety in our physical implementation. For example, if we wish to apply $\tau_{max}$ at the edge of the support polygon, we require a Cartesian stiffness of $2810$ N/m. 

In addition to the joint controllers, the gravity compensation term $m_{tot}g\hat{z}$ and upward assistive force $F_{assist}$ are included in the total desired open-loop force production $F_0$ in the body frame before passing through $\mathbb{J}^T$. 

We assume that the production of the required actuator torques is handled by the motor current controllers at a high enough bandwidth that we do not need to take their dynamics into account.

\section{Implementation and Validation}\label{secImplExp}
\begin{figure*}[t]
	\centering
	\includegraphics[width=\textwidth]{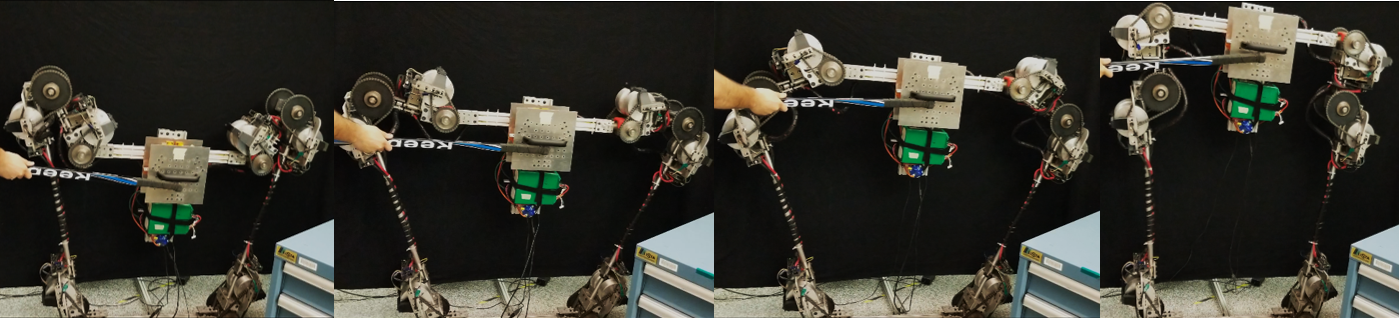}
	\caption{Sequence of the prototype XRL System being led through a squatting motion}
	\label{figSquatSequence}
\end{figure*}

The XRL System Prototype (See Fig. \ref{figXRLWorn} and Fig. \ref{figSquatSequence}) was designed to meet the force requirements of the squatting and crouching task while remaining in kinematically efficient configurations during operation \cite{Gonzalez2018}. It consists of two legs with three 2-DOF joint modules each, one at the hip, one at the knee, and one at the ankle. Each joint module consists of two brushless motors connected through a chain drive reduction to a differential mechanism. The differential couples the motor pair output and allows both motors to apply their maximum force along the same dimension, effectively doubling the force output without requiring more gearing. 

Each pair of motors at each joint module is controlled by the 24V version of the ODrive open-source brushless motor controller \cite{Weigl2018}. By measuring the motor position with an AMS AS5147P magnetic absolute encoder, these controllers use Field Oriented Control to drive current to the torque-producing axes of the motors, resulting in continuous torque production. Each of the 6 ODrives are connected to a computer via USB, and a Python program calculates and sends the desired trajectory reference, control gains, and open-loop torques using angle feedback from the controllers. For this demonstration, a handle on the operator side of the sensor is used to manipulate the XRL system from a safe distance with the end of a hockey stick.
\begin{figure}[t]
	\centering
	\includegraphics[width=0.825\linewidth]{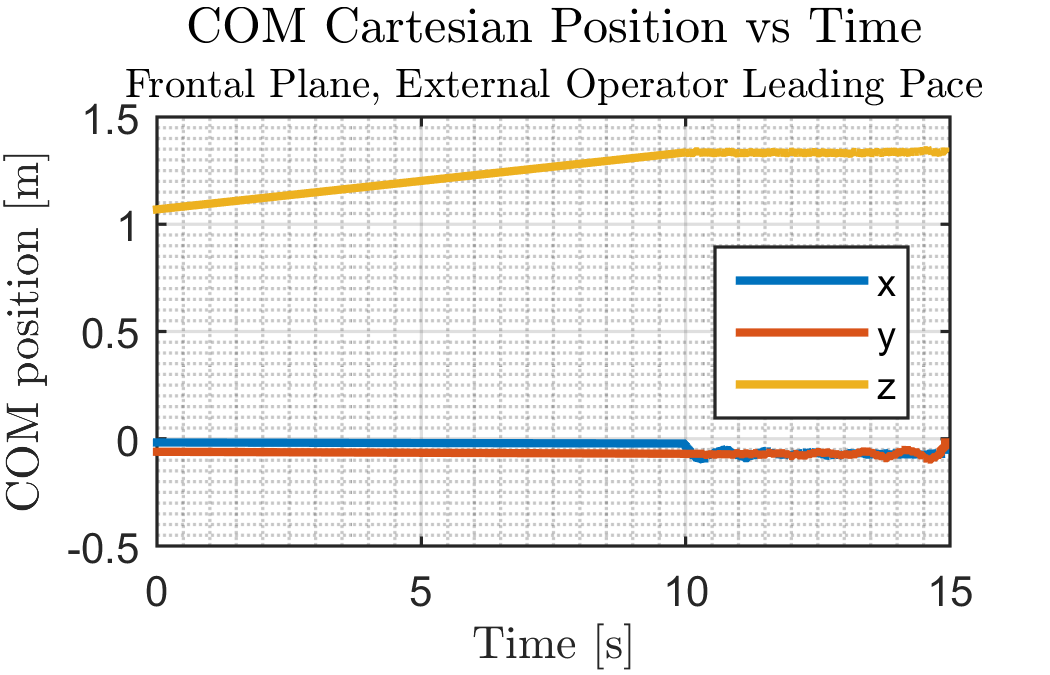}
	\caption{Cartesian XRL body position vs time while an external operator leads the pace of squatting motion during Hybrid Open-Loop Closed-Loop Control. At 10 seconds, the system undergoes a simulated disconnection from the centralized controller, but maintains stability despite destabilization attempts.}
	\label{figResults}
\end{figure}

To verify the control architecture, a human operator led the XRL System through the motions of a squat in the vertical direction while it balanced and stabilized itself in all other directions. Fig. \ref{figResults} shows the measured XRL trajectory in Cartesian space over time. As can be seen by the irregular trajectory from 0 seconds to 10 seconds, the robot balances itself in $X$ and $Y$ while the human leads the pace of motion in $Z$. A simulated communication failure occurs at 10 seconds, demonstrating the robot's ability to gracefully maintain its balance thanks to the distributed control architecture. As can be seen in Fig. \ref{figSquatSequence}, the operator leads the pace as the XRL maintains balance and stability throughout the squatting motion.

\section{Conclusion and Future Work}\label{secConclusion}
The Extra Robotic Legs System aims to empower the industrial worker and enhance their ability to perform their job by alleviating the burden of heavy equipment and enabling them to execute strenuous maneuvers more easily. The XRL completely bears the equipment load, allowing the unhindered operator to perform their tasks, enabled by a kinematic structure that is independent of the human's configuration. The robot balances itself and maintains a stable posture while simultaneously allowing the human operator the agency to lead a transition in posture from standing to crawling. While transitioning and in the crawling configuration, the XRL applies an upward assistive load to the human operator, helping them hold their posture while performing their task with both hands, crouch down, and stand back up. 
A Hybrid Open-Loop / Closed-Loop Control Architecture was developed to mix the position feedback control and force feedforward modes in a systematic manner to individual joint feedback control.
The control framework has been successfully tested on our XRL System Prototype. 

Next steps for the XRL System include testing with a human user wearing the 6-DOF force-torque sensing harness, stability analysis and implementation of dynamic balance control during walking, synchronizing the walking gait with the human operator using a sensing suite to observe and react to their movements, and finalizing the intent recognition framework to allow seamless transitions between the various modes of motion. The XRL System will eventually be tested with nuclear facility decontamination personnel, with the intent of further developing the system to be deployed at U.S. Department of Energy locations. 

Future applications beyond the DOE environmental management mission include bearing the weight of a first responder's loadout during natural disasters, relieving recreational hikers of their gear load as they maneuver, assisting construction and other industrial workers who must take uncomfortable postures near the ground for extended periods, and serving in a rehabilitation setting as ``training wheels'' for patients recovering from a walking impairment. 
	
\bibliographystyle{IEEEtran}
\bibliography{IEEEabrv,mybib}

\begin{thebibliography}{10}
\providecommand{\url}[1]{#1}
\csname url@samestyle\endcsname
\providecommand{\newblock}{\relax}
\providecommand{\bibinfo}[2]{#2}
\providecommand{\BIBentrySTDinterwordspacing}{\spaceskip=0pt\relax}
\providecommand{\BIBentryALTinterwordstretchfactor}{4}
\providecommand{\BIBentryALTinterwordspacing}{\spaceskip=\fontdimen2\font plus
\BIBentryALTinterwordstretchfactor\fontdimen3\font minus
  \fontdimen4\font\relax}
\providecommand{\BIBforeignlanguage}[2]{{%
\expandafter\ifx\csname l@#1\endcsname\relax
\typeout{** WARNING: IEEEtran.bst: No hyphenation pattern has been}%
\typeout{** loaded for the language `#1'. Using the pattern for}%
\typeout{** the default language instead.}%
\else
\language=\csname l@#1\endcsname
\fi
#2}}
\providecommand{\BIBdecl}{\relax}
\BIBdecl

\bibitem{NIOSH1985}
NIOSH, OSHA, USCG, and EPA, \emph{{Occupational Safety and Health Guidence
  Manual for Hazardous waste Site Activities}}, Washington, DC, 1985.

\bibitem{BureauofLaborStatistics2015a}
{Bureau of Labor Statistics}, ``{Nonfatal occupational injuries and illnesses
  requiring days away from work},'' US Department of Labor, Tech. Rep.
  USDL-15-2205, 2015.

\bibitem{CalgaryRegionalHealthAuthority2010}
C.~C. {Calgary Regional Health Authority}, ``{The Direct and Indirect Costs of
  Workplace Injuries},'' Encompass Group, LLC, Tech. Rep.~3, 2010.

\bibitem{Kurek2017}
D.~A. Kurek and H.~H. Asada, ``{The MantisBot: Design and impedance control of
  supernumerary robotic limbs for near-ground work},'' \emph{Proceedings - IEEE
  International Conference on Robotics and Automation}, pp. 5942--5947, 2017.

\bibitem{Gonzalez2018}
D.~J. Gonzalez and H.~H. Asada, ``{Design of Extra Robotic Legs for Augmenting
  Human Payload Capabilities by Exploiting Singularity and Torque
  Redistribution},'' \emph{IEEE International Conference on Intelligent Robots
  and Systems (IROS)}, 2018.

\bibitem{Mulder2015}
M.~Mulder, D.~A. Abbink, and T.~Carlson, ``{Introduction to the Special Issue
  on Shared Control: Applications},'' \emph{Journal of Human-Robot
  Interaction}, vol.~4, no.~3, p.~1, 2015.

\bibitem{Raibert1981}
M.~H. Raibert and J.~J. Craig, ``{Hybrid Position/Force Control of
  Manipulators},'' \emph{Journal of Dynamic Systems, Measurement, and Control},
  vol. 103, no.~2, p. 126, 1981.

\bibitem{Fisher1992}
W.~D. Fisher and M.~S. Mujtaba, ``{Hybrid Position/Force Control: A Correct
  Formulation},'' \emph{International Journal of Robotics Research}, vol.~11,
  no.~4, pp. 299--311, 1992.

\bibitem{Hogan1984}
N.~Hogan, ``{Impedance Control: An Approach to Manipulation},'' \emph{IEEE
  American Control Conference}, pp. 304--313, 1984.

\bibitem{Seraji1994}
H.~Seraji, ``{Adaptive admittance control: an approach to explicit force
  control in compliant motion},'' \emph{Proceedings of the 1994 IEEE
  International Conference on Robotics and Automation}, pp. 2705--2712, 1994.

\bibitem{Schimmels1992}
J.~M. Schimmels and M.~A. Peshkin, ``{Admittance Matrix Design for Force-Guided
  Assembly},'' \emph{IEEE Transactions on Robotics and Automation}, vol.~8,
  no.~2, pp. 213--227, 1992.

\bibitem{Ficuciello2015}
F.~Ficuciello, L.~Villani, and B.~Siciliano, ``{Variable Impedance Control of
  Redundant Manipulators for Intuitive Human-Robot Physical Interaction},''
  \emph{IEEE Transactions on Robotics}, vol.~31, no.~4, pp. 850--863, 2015.

\bibitem{Albu-Schaffer2002}
A.~Albu-Sch{\"{a}}ffer and G.~Hirzinger, ``{Cartesian Impedance Control
  Techniques for Torque Controlled Light-Weight Robots},'' in \emph{IEEE
  International Conference on Robotics and Automation (ICRA)}, no. May, 2002,
  pp. 657--663.

\bibitem{Ruiz2017}
M.~R. Ruiz, ``{Design and Analysis of a Stewart-Platform-Based Six-Axis Load
  Cell},'' S.B. Thesis, Massachusetts Institute of Technology, 2017.

\bibitem{Guizzo2015}
E.~Guizzo, E.~Ackerman, and F.~Shi, ``{DARPA Robotics Challenge: Amazing
  Moments, Lessons Learned, and What's Next},'' \emph{IEEE Spectrum}, 2015.

\bibitem{Atkeson2015}
C.~G. Atkeson, B.~P. Babu, N.~Banerjee, D.~Berenson, C.~P. Bove, X.~Cui,
  M.~DeDonato, R.~Du, S.~Feng, P.~Franklin, M.~Gennert, J.~P. Graff, P.~He,
  A.~Jaeger, J.~Kim, K.~Knoedler, L.~Li, C.~Liu, X.~Long, T.~Padir, F.~Polido,
  G.~G. Tighe, and X.~Xinjilefu, ``{No falls, no resets: Reliable humanoid
  behavior in the DARPA robotics challenge},'' \emph{IEEE-RAS International
  Conference on Humanoid Robots}, vol. 2015-Decem, pp. 623--630, 2015.

\bibitem{VanderVorm2015}
J.~{Van der Vorm}, L.~O'Sullivan, R.~Nugent, and M.~de~Looze, ``{Considerations
  for developing safety standards for industrial exoskeletons.}''
  \emph{Robo-Mate}, no. May, pp. 1--13, 2015.

\bibitem{Roderick2007}
S.~Roderick and C.~Carig, ``{Designing Safety-Critical Rehabilitation
  Robots},'' \emph{Rehabilitation Robotics}, no. August, 2007.

\bibitem{Hogan1990}
N.~Hogan, ``{Mechanical Impedance of Single- and Multi-Articular Systems},'' in
  \emph{Multiple Muscle Systems: Biomechanics and Movement Organization}.\hskip
  1em plus 0.5em minus 0.4em\relax New York: Springer-Verlag, 1990, ch.~9.

\bibitem{Chen1999}
S.~Chen and I.~Kao, ``{Theory of Stiffness Control in Robotics Using the
  Conservative Congruence Transformation},'' \emph{Int. Symp. of Robotics
  Research}, pp. 7--14, 1999.

\bibitem{Weigl2018}
\BIBentryALTinterwordspacing
O.~Weigl, ``{ODrive: High Performance Motor Control},'' 2018. [Online].
  Available: \url{https://odriverobotics.com/}
\BIBentrySTDinterwordspacing

\end{thebibliography}
	

\end{document}